\documentclass{article}

\usepackage{arxiv}

\usepackage[utf8]{inputenc} 
\usepackage[T1]{fontenc}    
\usepackage{hyperref}       
\usepackage{url}            
\usepackage{booktabs}       
\usepackage{amsfonts}       
\usepackage{nicefrac}       
\usepackage{microtype}      
\usepackage{lipsum}		
\usepackage{graphicx}
\usepackage{natbib}
\usepackage{doi}

\usepackage[linesnumbered,ruled,vlined]{algorithm2e}
\usepackage{mathtools}

\usepackage{amsmath,amssymb,amsfonts}
\usepackage{algorithmic}
\usepackage{graphicx}
\usepackage{textcomp}
\usepackage{xcolor}
\usepackage{amsmath,bm}
\usepackage{url}
\usepackage{multirow}
\usepackage{xcolor,colortbl}

\definecolor{Gray}{gray}{0.85}
\definecolor{LightCyan}{rgb}{0.80,0.90,0.75}
\definecolor{LightBlue}{rgb}{0.67,0.84,0.90}

\usepackage{balance}
\usepackage[ruled,vlined,linesnumbered]{algorithm2e}
\usepackage{comment}

\usepackage{subfig}
\usepackage{mwe}
\usepackage{float}
\usepackage{listings}
\usepackage{enumitem}
\usepackage{textcomp}
\usepackage{caption}
\usepackage[utf8]{inputenc}

\def\BibTeX{{\rm B\kern-.05em{\sc i\kern-.025em b}\kern-.08em
    T\kern-.1667em\lower.7ex\hbox{E}\kern-.125emX}}

\title{Toward efficient resource utilization at edge nodes in federated learning}

\author{Sadi Alawadi\\
        Department of Computer Science \\
Blekinge Institute of Technology, Karlskrona, Sweden\\
        \texttt{sadi.alawadi@bth.se \thanks{Corresponding author.}}
        \And
        Addi Ait-Mlouk \\
        School of Informatics, University of Sk\"ovde,\\
        Sk\"ovde, Sweden\\
        \texttt{addi.ait-mlouk@his.se }
         \And
        Salman Toor\\
        Department of Information Technology\\
        Division of Scientific Computing\\
        Uppsala University, Sweden\\
        Scaleout Systems, Sweden\\
        \texttt{salman.toor@it.uu.se}
         \And
         Andreas Hellander \\
        Department of Information Technology\\
        Division of Scientific Computing\\
        Uppsala University, Sweden\\
        Scaleout Systems, Sweden\\
        \texttt{andreas.hellander@it.uu.se}
}

\renewcommand{\shorttitle}{Toward efficient resource utilization at edge nodes in federated learning}

\hypersetup{
pdftitle={A template for the arxiv style},
pdfsubject={q-bio.NC, q-bio.QM},
pdfauthor={David S.~Hippocampus, Elias D.~Striatum},
pdfkeywords={First keyword, Second keyword, More},
}

\begin{document}
\maketitle

\begin{abstract}
Federated learning (FL) enables edge nodes to collaboratively contribute to constructing a global model without sharing their data. This is accomplished by devices computing local, private model updates that are then aggregated by a server. However, computational resource constraints and network communication can become a severe bottleneck for larger model sizes typical for deep learning (DL) applications. Edge nodes tend to have limited hardware resources (RAM, CPU), and the network bandwidth and reliability at the edge is a concern for scaling federated fleet applications. In this paper, we propose and evaluate a FL strategy inspired by transfer learning in order to reduce resource utilization on devices, as well as the load on the server and network in each global training round. For each local model update, we randomly select layers to train, freezing the remaining part of the model. In doing so, we can reduce both server load and communication costs per round by excluding all untrained layer weights from being transferred to the server.  The goal of this study is to empirically explore the potential trade-off between resource utilization on devices and global model convergence under the proposed strategy. We implement the approach using the FL framework FEDn. A number of experiments were carried out over different datasets (CIFAR-10, CASA, and IMDB), performing different tasks using different DL model architectures. Our results show that training the model partially can accelerate the training process, efficiently utilizes resources on-device, and reduce the data transmission by around 75\% and 53\% when we train 25\%, and 50\% of the model layers, respectively, without harming the resulting global model accuracy. Furthermore, our results demonstrate a negative correlation between the number of participating clients in the training process and the number of layers that need to be trained on each client's side. As the number of clients increases, there is a decrease in the required number of layers. This observation highlights the potential of the approach, particularly in cross-device use cases.
\end{abstract}


\maketitle

\section{Introduction}\label{sec1}

Federated learning (FL) is a privacy-preserving machine learning (ML) training strategy introduced by McMahan et al. \cite{mcmahan2017communication}. 
In FL, edge nodes contribute to a global model by locally computing partial model updates, which are then exchanged with a server and combined/aggregated into a global model. By iterating this process, we avoid sharing or transferring private data \cite{konevcny2016federated, fedhealth, fedqas} instead of moving data to a central server, the model implementation is transferred to the data owners’ local sites, where model training occurs. In this sense, FL falls in the category of decentralized optimization.

In the most basic FL architecture, a single central server constructs a global model in each communication (training) round by aggregating model parameters sent by the edge nodes. The limited internet connection bandwidth makes the model weight transfer between the edge nodes and server a bottleneck, contributing significantly to each round's training time \cite{chen2021communication}. In addition, 
training a large and complex Neural Network (NN) model at the edge node requires non-trivial time and computational resources (memory, network and CPU). Taken together, it is a challenge to accommodate increasingly complex models in cases where the network connection and the edge hardware have limitations. 

FL often involves edge devices that are not homogeneous and have varying computational capabilities. As a result, the research community has focused on addressing the challenges posed by weak computing capacity devices, commonly referred to as stragglers \cite{stragglers1, stragglers3}. These devices can significantly slow the training process and may need to be excluded from the round. Consequently, training a complex model over massive data in federated settings is challenging due to the presence of stragglers, which can slow down the process and limit the node's contribution to the round \cite{stragglers2}.

Therefore, different approaches have been proposed that can reduce resource utilization during the training of a deep learning (DL) model, both in centralized settings and decentralized settings. One such strategy is to freeze pre-trained model layers and to add a new output layer to be trained over the new task data, where its weights will be adapted based on the weights of the prior layers (transfer learning or fine-tuning) \cite{too2019comparative, liu2021autofreeze}. This approach is applicable in a centralized paradigm. For distributed settings, two main parallel strategies are in use to speed up training: 1) replicating a copy of the whole model overall cluster nodes and then using data mini-batches broadcasted by the cluster head (data parallelism) \cite{chen2018efficient, zou2014mariana}, and 2) distributing the model's layers over the cluster nodes, where each layer will be trained using the entire data set, and a cluster coordinator is responsible for parallel communication (model parallelism) \cite{vishnu2016distributed,hewett2020linear,jia2018beyond, shoeybi2019megatron}.

Recently, a new approach inspired by transfer learning aims to selectively freeze a set of layers and only update the remaining layers in each iteration \cite{xiao2019fast}. In this paper, we explore the same strategy for local model training in the FL setting. The goal is to a) reduce the resource needs on the edge device and, in this way, fit larger models by reducing the training memory and CPU footprint for each update, and b) reduce network transfer costs by reducing the number of updated parameters in each iteration. We are specifically interested in investigating the feasibility of this approach for scenarios where the hardware on the device is too limited to train the entire model effectively (resource starvation). To summarize, the main contributions of this article are:

\begin{itemize}
\item Our proposed approach draws inspiration from transfer learning and aims to decrease the amount of resources used on edge nodes/clients. By doing so, we aim to enhance efficiency and dependability, thereby minimizing the occurrence of stragglers in a round. Additionally, this approach is intended to enable edge devices/nodes with constrained resources to operate more effectively and provide a stable training environment.
\item Reduce the amount of transferred data (gradients, weights) over the network between the edge nodes and the model aggregator.
\item Systematically evaluates the potential of the approach for varying resource availability on the client side, including the Jetson Nano \footnote{https://developer.nvidia.com/embedded/jetson-nano-developer-kit} as a constrained edge device. 
\end{itemize}

The rest of this paper is organized as follows. Section \ref{sec:relatedWork} provides background on FL, model fine-tuning, transfer learning and related work, as well as how the proposed approach relates to general parallel training strategies. Next, we introduce the proposed approach in Section \ref{sec:material}. Section \ref{sec:results} describes the experimental settings. Finally, we conclude the paper in Section \ref{sec:conc}.

\section{Background and Related Work} \label{sec:relatedWork}

\subsection{\textbf{Federated Learning}}
FL is a new solution proposed by Google \cite{konevcny2016federated, konevcny2016federated1,fedqas} to preserve data privacy that aliens with General Data Protection Regulation (GDPR). FL enables users or organizations to collaboratively train a ML model without transferring their own data to a central storage system. Instead, the code is moved to the data owners' local sites in such a paradigm; incremental local updates are combined into a global model every communication round \cite{alawadi2024fedcsd}. Typically, a server or aggregator is responsible for managing the client nodes, communication, distributing the global model weight, controlling the training rounds, and generating a global model from received models using FEDAvg\cite{mcmahan2017communication}, FEDProx \cite{sahu2018convergence} and many other aggregation methods. Firstly, the server starts the training by distributing the ML model architecture with random weight. Secondly, each client will receive the shared model, initiate the training process using local data, and share the weight updates with the server. Third, in each training round, the server combines all obtained weight updates using FEDAvg, or other aggregation methods to update the global model weight. Finally, new weights are distributed across the clients to start a new training round; in this sense, all clients have shared their knowledge without sharing their data \cite{ekmefjord2021scalable}.
Hence, FL maintains data privacy, reduces data transferring costs, and shares different client or organization knowledge, especially in healthcare scenarios where the data is more sensitive. In addition, learning from different sources is required to capture more knowledge about a given problem.\cite{yang2019federated}.

\subsection{\textbf{Transfer Learning and Model Fine-tuning}}
The shortage of the data samples needed to train a ML model enables transfer learning and fine-tuning to reduce the training cost through transferring the knowledge from a large pre-trained model (source) to 
a new model (destination) using the small number of new samples that will perform new tasks \cite{guo2019spottune, vrbanvcivc2020transfer}.

In transfer learning, only the last few layers of a large source model will be trained over a new dataset for a specific task and adapt their weights based on the prior layers \cite{vrbanvcivc2020transfer}. While in the fine-tuning case, the whole pre-trained model is trained over a new dataset and new task \cite{too2019comparative}. Despite their difference, both techniques share the common aim of enabling the use of DL in situations where there is a dearth of training data. This is achieved by transferring knowledge across related domains, substantially reducing training time and resource consumption in a centralized fashion.

\subsection{\textbf{Training Parallelization Techniques}}

Here, we discuss the two main approaches that have been introduced to address challenges in large-scale model training. In both cases, the focus is on parallelizing the training process by utilizing large computational resources.

\begin{enumerate}

\item Training DL models requires a massive amount of data and computational resources, which can be time-consuming. The data parallelism  \cite{valiant1990bridging} approach has been widely used to address this challenge, using a parameter server architecture to distribute the training workload across multiple workers and speed up the process. In this approach, the training dataset is divided into mini-batches and each worker is assigned a different subset. The parameter server maintains a full copy of the DL model and communicates with workers to synchronize gradients or weights \cite{shallue2018measuring}. Each worker receives a copy of the DL model and a mini-batch of data, computes local gradients, and shares them with the server parameter. The server then updates the model parameters based on the gradients received from the workers, using negative gradient direction or parameter averaging. The latest values are shared with the workers \cite{park2020hetpipe, shazeer2018mesh}. However, the communication channel bandwidth capacity can become a bottleneck as more workers join the training process, leading to slower processing times. To address this issue, a new architecture called AllReduce operations was proposed, which does not depend on the number of workers in the training process to maintain communication channel capacity \cite{thakur2005optimization}.

\item To tackle complex tasks that require large DL models with millions of parameters, training these models can be very computationally demanding and resource-intensive. Model parallelism is a useful approach to efficiently train these models by creating a cluster of worker nodes with a coordinator to parallelize the training process \cite{dean2012large}. The DL model is split into sub-layers and distributed among multiple workers, with the coordinator maintaining the model layers in sequential order, managing data flow, and communicating between all workers. Each worker is assigned a mini-batch of data that is shared among all workers to update the worker-assigned model layers' gradients. Finally, the coordinator combines all the layers received from workers to produce the final model. This approach can be implemented in multi-GPU cores, where each core acts as an independent worker \cite{krizhevsky2014one, park2020hetpipe}.
\end{enumerate}

\subsection{Related Work}
Several studies have been investigated speeding up and reducing the cost of training DL models. These studies are oriented to solve different problems before and during the training process, such as complex and large models, shortage in the training sample, a vast amount of training data that demand high computational resources and time. Transfer Learning has been proposed to tackle lack of training data via transferring the knowledge from a related pre-trained model to a new task \cite{pan2009survey} which is widely used in image processing and natural language processing (NLP). 

Different approaches have been proposed to distribute the training process workload across a group of machines (Cluster). Dean et al. \cite{dean2012large} have developed a DistBelief framework that supports DL model parallelism for large models. The framework trains large models over a computing cluster with thousands of machines. It comprises two main algorithms: Downpour SGD, responsible for a large number of model replicas and adaptive learning rates, while Sandblaster is accountable for the parallelization process. In this study, \cite{jia2018beyond} authors have investigated the space of parallelization strategies (i.e. SOAP), which includes Sample, Operation, Attribute, and Parameter dimensions to parallelize a DL model. Furthermore, they proposed FlexFlow framework \cite{jia2018beyond} that uses the SOAP space to search randomly for a fast parallelization strategy for a specific machine. In addition to the frameworks mentioned above, more research has been conducted on model parallelism, such as \cite{sergeev2018horovod, park2020hetpipe, gaunt2017ampnet,chen2018efficient}.

For a large training data sample, Valiant \cite {valiant1990bridging} has introduced the bulk-synchronous parallel (BSP) model, which parallelises the training process by two main steps. Firstly, a replica of the whole DL model will be placed on each device. Secondly, the training dataset will be split into mini-batches and then distributed among multiple workers to train the DL model.
Finally, each worker synchronises model parameters with a different worker at the end of each iteration \cite{krizhevsky2012imagenet}. Moreover, Tensorflow \cite{tensorflow}, Coffe2 \cite{caffe2}, and Pytorch \cite{pytorch} frameworks have been used in both data and model parallelism to parallelise the DL training process. Nevertheless, Data parallelism is an efficient technique that can train a small DL model with few parameters. While in a large model's case, this becomes an inefficient strategy that causes a scalability jam in large-scale distributed training environments.

The authors \cite{freezeout} proposed a Freezeout approach to accelerate the training process by training each hidden layer in the model for a set part of the training schedule, "freezing out layers" progressively and the back-propagation of these layers is avoided. Also, Chen et al. \cite{xu2019notice} have followed the same strategy proposed in \cite{freezeout}  to train the model by freezing the hidden layers out one by one.
A new approach has been proposed by Xueli et al. \cite{fastdl} to freeze layers intelligently during the training phase, where the differences of the normalized gradient for all weighted layers have been computed to identify the number of layers that should be frozen. This approach has been developed on top of stochastic gradient descent (SGD) and evaluated using large models (i.e. VGG, ResNets, and DenseNets) in a centralized fashion.   

Identifying the number of the freezing layers has been investigated in \cite{lee2019would} during the transformer fine-tuning process for well know pre-trained models (i.e. BERT, and RoBERTa) in the NLP field. Moreover, Yuhan et al.\cite{autofreeze} were proposed AutoFreeze framework 
for automatically freezing layers to speed up fine-tuning by applying an adaptive approach to identify all layers that need to be trained while maintaining accuracy. Also,  multiple mechanisms have been developed to decrease the forward computation time while conducting model fine-tuning by enabling client caching of intermediate activation's. 

Based on the fact that the internal layer's training progress differ significantly, a knowledge-guided training system (KGT) \cite{wang2022efficient} has been proposed to focus more on those layers. Sub consequently, KGT skips part of the computations and communications in the deep neural network's (DNN) internal layers (hidden layers) to accelerate the training process while maintaining accuracy. 

Chen et al. \cite{chen2021communication} were proposed a new scheme named Adaptive Parameter
Freezing (APF) 
tackle the communication bottleneck in FL settings. The APF is responsible for freezing and unfreezing converged parameters during the training rounds for intermittent periods. The model was fully trained to identify the stable and unstable parameters for several rounds, and then APF freezes the stable parameters based on threshold increase gradually. In addition, the authors have introduced a mechanism that dynamically adjusts the stability threshold at runtime when most of the parameters have been classified as stable and decreases the stability threshold by one-half. The results reveal that this scheme has reduced the communication volume without compromising the model's accuracy. However, this approach still relies on memory to cash the prior parameters to check their stability and also requires both CPUs and RAMs to train the entire model at the beginning, and each unfreezes period. While our approach train sub-layers of the model selected randomly, which significantly impacts the resources, training time, transferred data, and the final model accuracy, as shown in our results. 

Our approach stands out from existing methods in this context due to its unique training strategy in FL settings. Our strategy focuses on training different parts of the model every communication round, aiming to achieve several key goals. These goals include, \textit{reducing resource utilization, minimizing the communication flow, and enabling the 
restricted edge devices to participate in the training process without sharing the raw data}. Unlike other approaches reported in the literature, where a complete model must be shared to train the model, our approach avoids the need to share a complete model. Instead, we adopt a more efficient technique where each client randomly selects a different model layer every communication round to be trained, which enables stragglers (constrained devices) to effectively participate in each round of the federated training.

\section{Proposed Approach} \label{sec:material}

\begin{figure}[!h]
\centering
\includegraphics[width=0.9\textwidth]{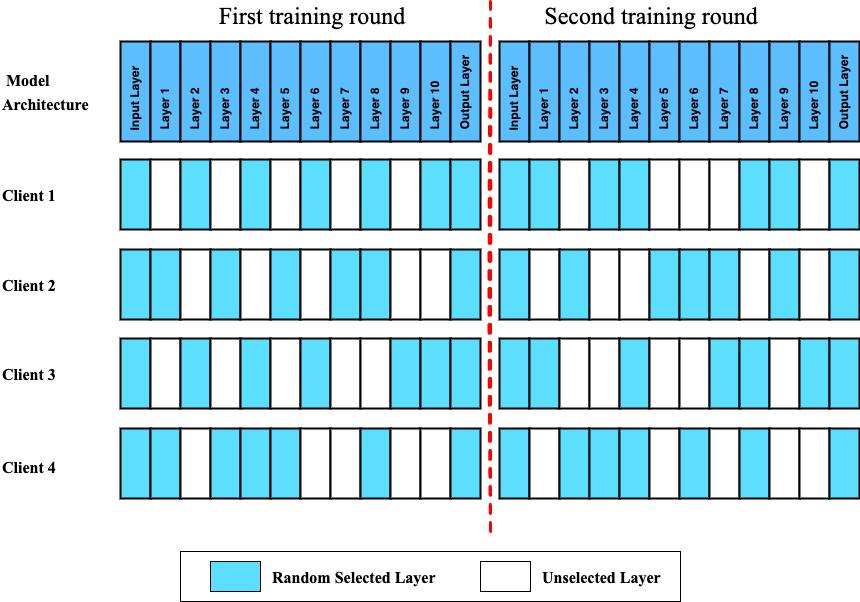}
\caption{The abstract diagram depicts the proposed approach for training the ML model with four clients in the FL context. Where each client independently selects 50\% of the entire model layers randomly during every training round.}
\label{fig:layerselection}
\centering
\end{figure}

The approach we propose involves selectively freezing layers during client updates. Figure \ref{fig:layerselection} shows the abstract diagram of our approach, where each client (i.e. four clients) independently selects a portion of the entire model randomly based on the identified percentage (50\% in this diagram, equivalent to five layers out of 10), determines the layers to be trained in FL settings during each communication round. As indicated in the diagram, it's important to note that each client trains different layers in every round.
This technique can be adapted to work with any federated training strategy or aggregation scheme, and it only requires modifications to be made on the client side. As a result, it offers a high degree of flexibility and versatility, making it easy to integrate into existing FL workflows.
While there are many different aggregation schemes with different properties, we have chosen to exemplify with the Federated Averaging (FedAvg) strategy in this work. FedAvg is an established FL aggregation algorithm proposed in \cite{mcmahan2017communication}.

The FedAvg algorithm is a decentralized version of SGD. In each training round, a subset of $C$ clients receive a copy of the recent global model. These clients execute $E$ epochs over their local dataset $D_k$ to update the model weights $W_k$. 
A new global model is constructed once the updated weights $W_k$ have been computed and sent back to the server. This new global model is built by averaging all the received updated weights $W_k$ contributed by all participating clients $n$ every round.

This weighted average is then utilized to generate a new version of the global model $M$. This process is repeated until the model converges. The FedAvg algorithm, which is unchanged in this paper, is outlined in Algorithm \ref{fedavg}. 
\begin{align}
\label{eq:averaging}
M = \sum_{k=1}^{C} \frac{n_k}{n}W_k^{(i)}.  
\end{align}

Each iteration on the client side is outlined in Algorithm \ref{client-update}. The only difference from standard FedAvg is the layer selection step (line 3). In our approach, a random layer selection strategy is used, although future work could explore more advanced selection strategies.


\begin{algorithm}
\DontPrintSemicolon
\SetAlgoLined
\KwIn{$W_t$}
\KwOut{$M(W_t)$}
\BlankLine
\textbf{Server executes:}\\
initialized \textbf{$W_0$}

\SetKwFunction{Server}{FedAVG}
\SetKwProg{Fn}{Function}{:}{}
\Fn{\Server{$k, W_{t-1}, W_t$}}{

\ForEach{$t\leftarrow 1$ \KwTo $r$}{

$S_t \leftarrow$ (sample a random set of clients)\\
\ForEach{client $k \in S_t$ \textbf{in parallel}}
{
$W_{t}^k \leftarrow ClientUpdate(k, W_t,N_l)$\;
$W_{t} \leftarrow \sum_{k=1}^{K} \frac{n_k}{n} W_{t}^k$\
}
$W_t \leftarrow (W_{t-1} + (W_{t}-W_{t-1})/t)$\
}
\Return $M(W_t)$

}
\caption{FedAVG algorithm. \textbf{C}: Number of clients, \textbf{r}: Number of rounds, \bm{$W_i$}: Local model weights and \textbf{M}: Global model weights}
\label{fedavg}
\end{algorithm}


\begin{algorithm}[!h]
\DontPrintSemicolon
\SetAlgoLined
\KwOut{\textbf{$W_t$}}

\SetKwFunction{client}{ClientUpdate} // Run on client k\\
\SetKwProg{Fn}{Function}{:}{}
\Fn{\client{$k, W_t,N_l$}}{
\textbf{$W_t(l)$} $ \leftarrow$ Select \textbf{$N_l$} layers from full model \textbf{$W_t$} randomly to train\\

\bm{$\beta$} $\leftarrow $ (split~$D^k$~into~mini~batches)\\
\For{$local~epoch~e_i \in 1, \dots e$}{
\For{batch~b $\in$ \bm{$\beta$}}{
$W_t \leftarrow W_t(l) - \eta \nabla l(W_t(l),b)$\;
}
}
\KwRet{$W_t$}\;
}
\caption{Local client update, \textbf{k}: Number of clients, \bm{$N_l$} : Number of trained layers, \bm{$D^k$}: Client k local dataset, \textbf{e}: Number of local epochs, and \bm{$\eta$} is the learning rate }
\label{client-update}
\end{algorithm}

In all experiments that follow we have used the FedAvg implementation in the FEDn FL framework \cite{ekmefjord2021scalable}. FEDn is highly scalable and fully distributed and can be used without modification for the evaluated strategy. This also illustrates that the approach can be embedded in a production-grade distributed system in a straightforward manner to reduce resource utilization on the edge device. It should be noted, however, that to fully benefit from the potential network transfer reduction, some minor modifications to the FEDn aggregation server would be needed.     

\section{Results and Discussion} \label{sec:results}

 Our carefully designed experiments have yielded valuable insights: in a federated training setting with restricted resources, choosing a limited number of layers at each client location can achieve results comparable to training the complete model at each site. A noteworthy observation we made was that a \textit{negative correlation} exists between the number of clients and the number of layers that need to be trained at each client location. As the number of clients increases, the required number of layers trained per client decreases, indicating that this approach can be especially effective for cross-device use cases. These findings provide valuable insights into FL optimization and offer practical recommendations for limited resource edge device deployment.
 
We outline our experiments' datasets, models, and configurations in the following subsections. We also introduce the evaluation metrics we used to assess the performance of our models (subsection \ref{ExpSett}). Afterwards, we provide a comprehensive discussion of the results we obtained (subsection \ref{ResDisc}).

\subsection{Experimental Settings} \label{ExpSett}

To explore the viability of our proposed approach, we utilized three open-source datasets and their associated ML models initialized with random weight, specifically selected to represent a diverse range of application domains (computer vision, NLP, and Internet-of-Things). Our investigation focused on evaluating model convergence, network load and communication cost, impact on training time, and resource utilization.

\begin{itemize}

\item \textbf{Experiment 1}: This experiment focuses on a computer vision task that utilizes CIFAR-10 dataset\footnote{\url{http://www.cs.toronto.edu/~kriz/cifar.html}}. The dataset consists of $60,000$ colour images with dimensions of $32\times32$ pixels, grouped into $10$ classes, with $6,000$ samples per each. The dataset is divided into two subsets: $50,000$ training images and $10,000$ test images. In this experiment, we randomly generated n-client data from Cifar10, ensuring that each client held an equal number of samples and that the data were independent and identically distributed (IID). Also, we used the VGG16 model \cite{cifar} in this experiment. Table \ref{vgg_arch} shows the model architecture, including layer types, output dimensions, and the number of trainable parameters per layer. The model has a total of $14,736,714$ parameters and $14$ trainable layers, including the output layer. The generated model size is $53.5MB$. For more information, please refer to the client source code available on GitHub.\footnote{\url{https://github.com/saadiabadi/cifar_updated.git}}. 

\item \textbf{Experiment 2}: 
This experiment is centred around sentiment analysis tasks using the IMDB dataset v1.0\footnote{\url{https://ai.stanford.edu/~amaas/data/sentiment/}}. The dataset consists of $50,000$ reviews, with a maximum of $30$ reviews per movie, equally divided between positive and negative reviews. In \cite{imdb}, the dataset has been used for sentiment analysis tasks, where a ML model predicts whether a given review is positive or negative based on the review text. We used a DL model to predict the review decision for this task. Table \ref{NLP_model_arch} shows the detailed architecture of the model, including the layer types, dimensions and parameters used to construct and generate the initial model. It's important to highlight that the dataset has been randomly partitioned among n clients, where each client received the same amount of data and maintained the same class distribution, meaning all clients followed the IID distribution. Further technical details can be found on the client source code GitHub\footnote{\url{https://github.com/saadiabadi/IMDB_Example.git}}.

\item \textbf{Experiment 3}: In this experiment, the focus is on human activity recognition (HAR) using the CASA dataset\footnote{\url{https://archive.ics.uci.edu/ml/datasets/Human+Activity+Recognition+from+Continuous+Ambient+Sensor+Data}}. The dataset comprises $13,956,534$ patterns collected over two months from 30 homes using continuous ambient and PIR sensors. Each pattern consists of a set of $37$ features linked to different sensors distributed throughout the home, representing daily human activities such as sleep, eating, reading, and watching TV \cite{HAR}. It's important to highlight that each home's data represents an individual client in this setting; both the data size and the number of patterns varied among clients, resulting in a non-independent and identified distributed (Non-IID) distribution. The goal of this experiment is to classify the output into $10$ different daily activities for each user. We use a Long Short-Term Memory (LSTM) model with an input layer dimension of $(100, 1, 36)$, four dense layers, and one output layer. The model has $68,884$ trainable parameters distributed across $6$ trainable layers, and the compiled model size is $254KB$. For further technical details, please refer to the client source code available on GitHub\footnote{\url{https://github.com/saadiabadi/Casa_IoT_Example.git}}.  
\end{itemize}


\begin{table*}[!ht]
\scriptsize
\caption{The VGG16 model architecture details used in computer vision experiment.}
\label{vgg_arch}
\begin{center}
\scalebox{.9}{
\begin{tabular}{|l|l|l||l|l|l||l|l|l|}
\hline
Layer type            & Output dimension        & Param \# & Layer type            & Output dimension      & Param \# & Layer type             & Output dimension      & Param \# \\ \hline
conv2d                  & (32, 32, 64)  & 1792     & activation\_4           & (8, 8, 256) & 0        & activation\_9            & (4, 4, 512) & 0        \\ \hline
batch\_normalization    & (32, 32, 64)  & 256      & conv2d\_5               & (8, 8, 256) & 590080   & max\_pooling2d\_3        & (2, 2, 512) & 0        \\ \hline
activation              & (32, 32, 64)  & 0        & batch\_normalization\_5 & (8, 8, 256) & 1024     & conv2d\_10               & (2, 2, 512) & 2359808  \\ \hline
conv2d\_1               & (32, 32, 64)  & 36928    & activation\_5           & (8, 8, 256) & 0        & batch\_normalization\_10 & (2, 2, 512) & 2048     \\ \hline
batch\_normalization\_1 & (32, 32, 64)  & 256      & conv2d\_6               & (8, 8, 256) & 590080   & activation\_10           & (2, 2, 512) & 0        \\ \hline
activation              & (32, 32, 64)  & 0        & batch\_normalization\_6 & (8, 8, 256) & 1024     & conv2d\_11               & (2, 2, 512) & 2359808  \\ \hline
max\_pooling2d          & (16, 16, 64)  & 0        & activation\_6           & (8, 8, 256) & 0        & batch\_normalization\_11 & (2, 2, 512) & 2048     \\ \hline
conv2d\_2               & (16, 16, 128) & 73856    & max\_pooling2d\_2       & (4, 4, 256) & 0        & activation\_11           & (2, 2, 512) & 0        \\ \hline
batch\_normalization\_2 & (16, 16, 128) & 512      & conv2d\_7               & (4, 4, 512) & 1180160  & conv2d\_12               & (2, 2, 512) & 2359808  \\ \hline
activation\_2           & (16, 16, 128) & 0        & batch\_normalization\_7 & (4, 4, 512) & 2048     & batch\_normalization\_12 & (2, 2, 512) & 2048     \\ \hline
conv2d\_3               & (16, 16, 128) & 147584   & activation\_7           & (4, 4, 512) & 0        & activation\_12           & (2, 2, 512) & 0        \\ \hline
batch\_normalization\_3 & (16, 16, 128) & 512      & conv2d\_8               & (4, 4, 512) & 2359808  & max\_pooling2d\_4        & (1, 1, 512) & 0        \\ \hline
activation\_3           & (16, 16, 128) & 0        & batch\_normalization\_8 & (4, 4, 512) & 2048     & average\_pooling2d       & (1, 1, 512) & 0        \\ \hline
max\_pooling2d\_1       & (8, 8, 128)   & 0        & activation\_8           & (4, 4, 512) & 0        & flatten                  & (512)       & 0        \\ \hline
conv2d\_4               & (8, 8, 256)   & 295168   & conv2d\_9               & (4, 4, 512) & 2359808  & dense                    & (10)        & 5130     \\ \hline
batch\_normalization\_4 & (8, 8, 256)   & 1024     & batch\_normalization\_9 & (4, 4, 512) & 2048     &                          &                   &          \\ \hline

\end{tabular}
}
\end{center}
\end{table*}



\begin{table}[!ht]
\scriptsize
\caption{Model specification for the sentiments analysis architecture.}
\label{NLP_model_arch}
\begin{center}
\scalebox{.88}{
\begin{tabular}{|ll|}
\hline
\multicolumn{1}{|l|}{\textbf{Layers}}                 & \textbf{Parameters characteristic}                              \\ \hline
\multicolumn{1}{|l|}{\textbf{Embedding layer}}        & max features= 20000, maxlen= 100, embedding size= 128 \\ \hline
\multicolumn{1}{|l|}{\textbf{Convolutional layer}}    & kernel size=5, filters= 64, pool size= 4              \\ \hline
\multicolumn{1}{|l|}{\textbf{LSTM layer}}             & lstm outputsize= 70                                   \\ \hline
\multicolumn{1}{|l|}{\textbf{One output Dense layer}} & 2 outputs (positive or negative review)               \\ \hline
\multicolumn{2}{|l|}{\textbf{The generated model size = 10MB}}                                                \\ \hline
\end{tabular}
}
\end{center}
\end{table}


For all conducted experiments, we used one local epoch, batch size 32, learning rate 0.01, and ADAM optimizer as the local training parameters settings. Moreover, Python and TensorFlow were used to implement the models and local model updates. The experiments were performed on the Swedish OpenStack Infrastructure as-a Service, SNIC Science Cloud (SSC)\footnote{\url{https://cloud.snic.se/}} \cite{8109140}.

To evaluate model performance, we used the accuracy function provided by Tensorflow, whose return value that falls between (0, 100), as shown in Equation \ref{eq:accuracy}, as well as the loss function using categorical cross-entropy as shown in Equation \ref{eq:loss}.

\begin{equation}
 Accuracy= \frac{TP + TN}{TP +TN + FN + FP}*100\%
 \label{eq:accuracy}
\end{equation}

\noindent where $TP$, $TN$,$FP$, and $FN$  are the True Positives, True Negatives, False Positives and False Negatives respectively.
\begin{equation}
Loss= -\sum_{i=1}^{N} y_i.\log\hat{y}_i
\label{eq:loss}
\end{equation}

where $\hat{y}_i$ is the model prediction for \textit{i-th} pattern, $y_i$ represent the corresponding real value, and \textit{N} is the total number of samples.

\subsection{Results and discussion}\label{ResDisc}

The results are organized to evaluate various aspects of our approach: model performance, the impact of scaling the number of the edge nodes (clients) on the model performance, the possible minimal size of data exchange through the network (network workload), the trained layers distribution, and finally, the resources utilization efficiency using both VMs and actual device (Jetson Nano).

\subsubsection{Model performance}\label{sec:performance}
\begin{figure}[!h]
\centering
\includegraphics[width=0.9\textwidth]{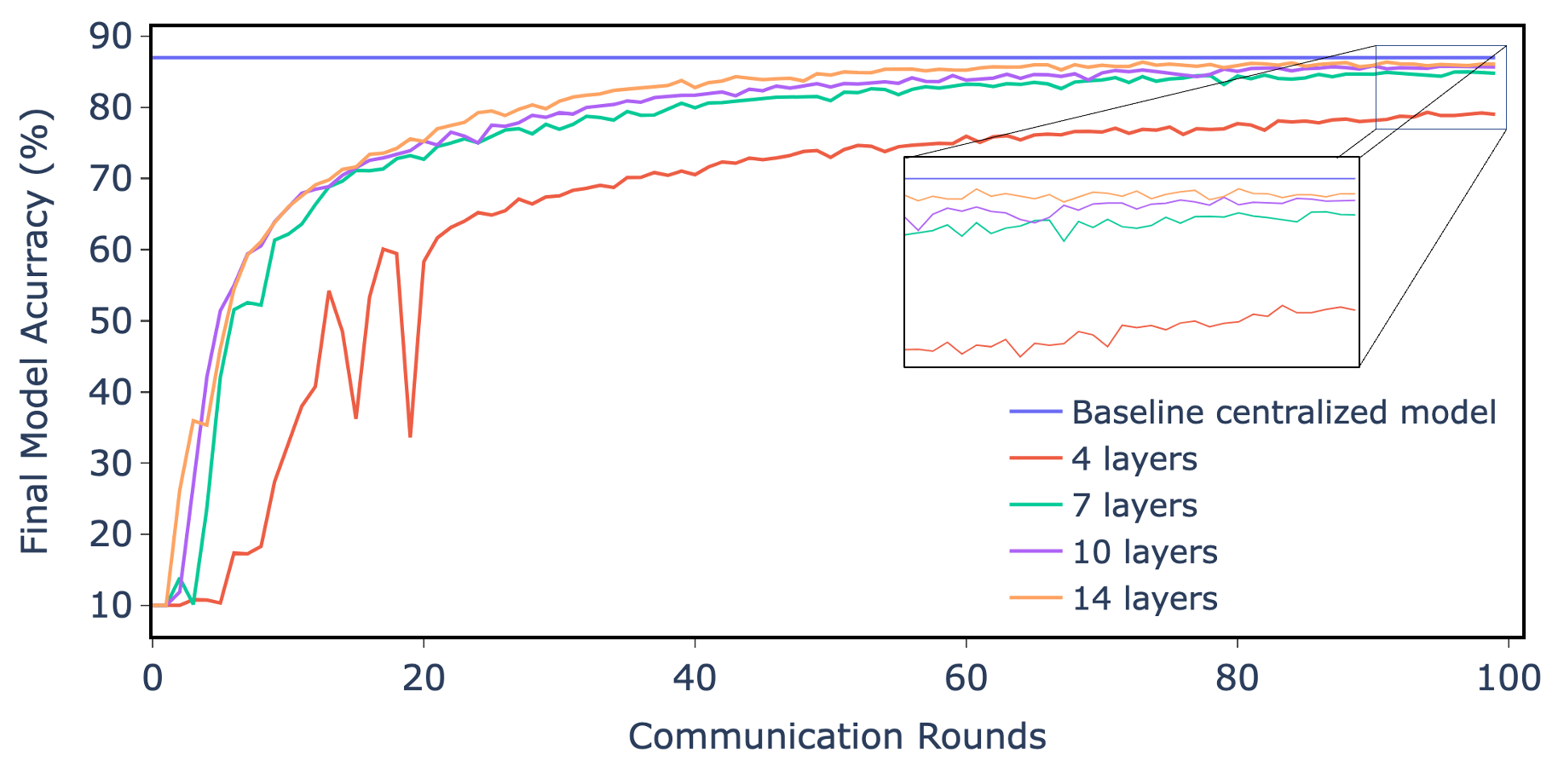}
\caption{VGG16 model accuracy for CIFAR-10 dataset using different numbers of trainable layers.}
\label{fig:modelAccuracy}
\centering
\end{figure}

 Our goal is to attain accuracy similar to conventional FL methods while implementing a new strategy that maximizes the utilization of edge resources. This approach not only upholds data privacy, which is crucial in any federated training environment but also minimizes training costs and reduces data transfer. To demonstrate the effectiveness of this strategy, we conducted a series of experiments where we trained different numbers of randomly selected model layers. Through these experiments, we systematically increased the number of layers to assess their impact on the overall performance of the model.
To compare the accuracy of centrally trained and FL global models, we varied the number of trainable layers for the FL settings. Figure~\ref{fig:modelAccuracy} shows the accuracy of the VGG16 model trained on the CIFAR-10 dataset, both centrally and in FL settings (10 clients and the data split partition was equally across 10 clients), with different trainable layers randomly selected per round.

The centralized model achieved an accuracy of $87.00\%$, while the FL model, with all layers included in the training process, achieved a slightly lower accuracy of $86.08\%$, a difference of only $0.92\%$. However, FL with slightly lower accuracy offers a privacy-preserving training environment.
Our experiments revealed that training $10$ randomly selected layers in each round of the model was enough to capture and learn the data behaviour, resulting around 85.70\% in accuracy. Figure~\ref{fig:modelAccuracy} indicates that the model began to converge from the first round of training. As we reduced the number of trained layers to approximately $50\%$ ($7$ layers) of the model, the accuracy gap increased. Nevertheless, the accuracy remained 84.79\% compared to the baseline, given the reduction in cost achieved through this approach, as elaborated in the subsequent sections.

 We also observed oscillations in the model's performance at the outset of training with only $4$ layers. The model faced difficulties in capturing the complete data behavior using the small number of layers in each round.
However, the model eventually achieves an accuracy of $79.00\%$ in the later stages, which is approximately $7\%$ lower than the baseline accuracy achieved through conventional federated training.

\begin{figure}[!htb]
    \centering
    \subfloat[Model accuracy trained over CASA dataset]{\includegraphics[width=0.95\textwidth]{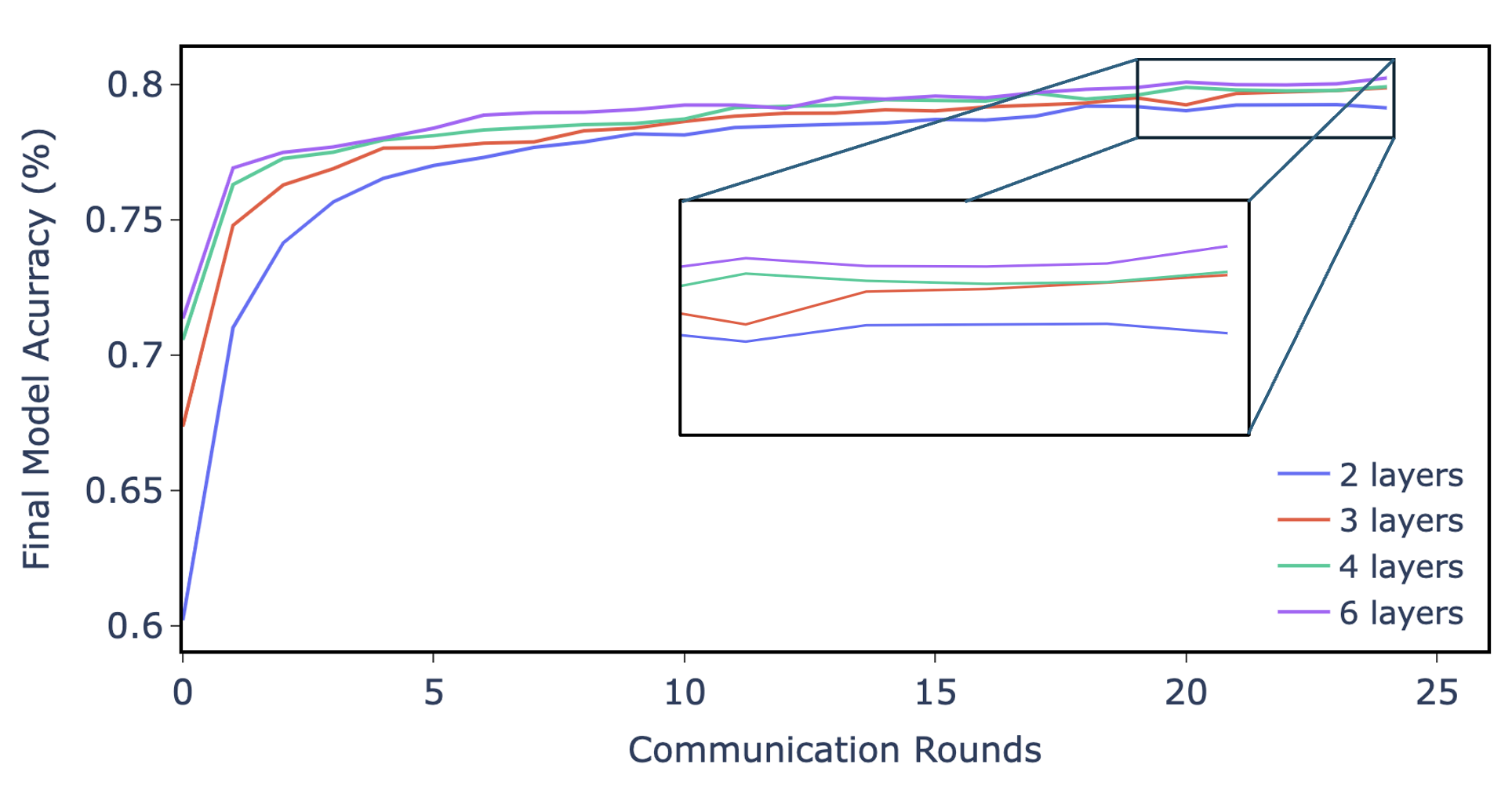}\label{fig:casa}}\\
    \subfloat[Model accuracy trained over IMDB dataset] {\includegraphics[width=0.9\textwidth]{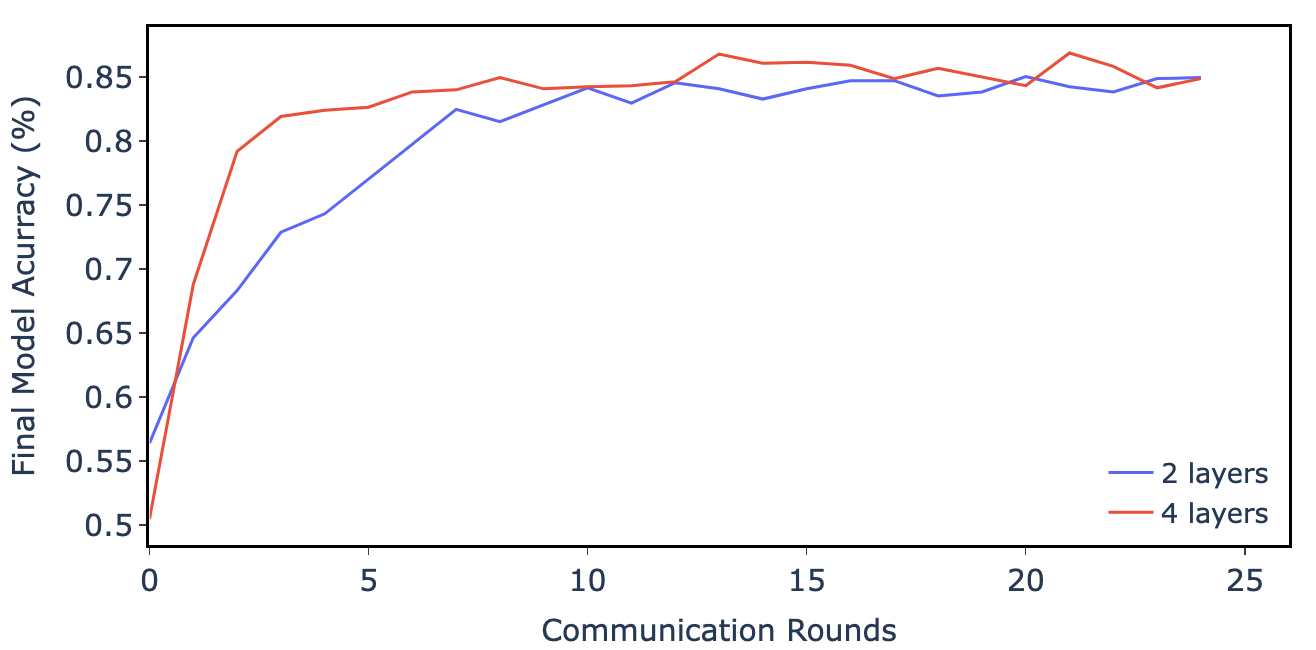}\label{fig:imdb}}
    \caption{Evaluating two different DL architectures to perform distinct tasks in terms of accuracy  (a) human activity recognition task using CASA dataset. (b) Sentiment analysis task using IMDB dataset}
    \label{fig:CasaIMDB}
\end{figure}
To highlight the advantages of our proposed approach, we evaluated its adaptability and robustness across two distinct domains, utilizing both CASA and IMDB datasets in FL settings, involving 10 clients, corresponding to 10 homes in the CASA dataset and equal subset of data amount split from the IMDB dataset. Figure~\ref{fig:CasaIMDB} illustrates a convergence in terms of model performance for both datasets. As depicted in Figure~\ref{fig:CasaIMDB}-a, for the HAR task, training only $33\%$ ($2$ randomly selected layers every round) of the model resulted in good accuracy (around 79.01\%) compared to the fully trained ($6$ layers, 80.20\%) model with a small gap. In contrast, training $50\%$ ($3$ layers) or $66\%$ ($4$ layers) of the model nearly achieved the same level of accuracy as training all the model layers. Interestingly, similar trends were observed in the NLP experiments, as demonstrated in Figure~\ref{fig:CasaIMDB}-b.

\subsubsection{Trainable layer distribution}
\begin{figure*}[!htb]
    \centering
    \subfloat[Train 25\% (4 Layers) of the model ]{\includegraphics[width=0.55\textwidth]{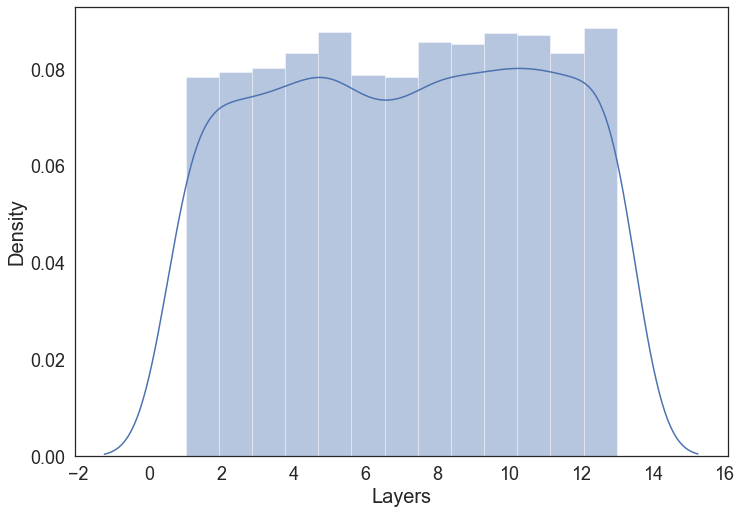}\label{fig:4l_distribution}}
    \subfloat[Train 50\% (7 Layers) of the model ] {\includegraphics[width=0.55\textwidth]{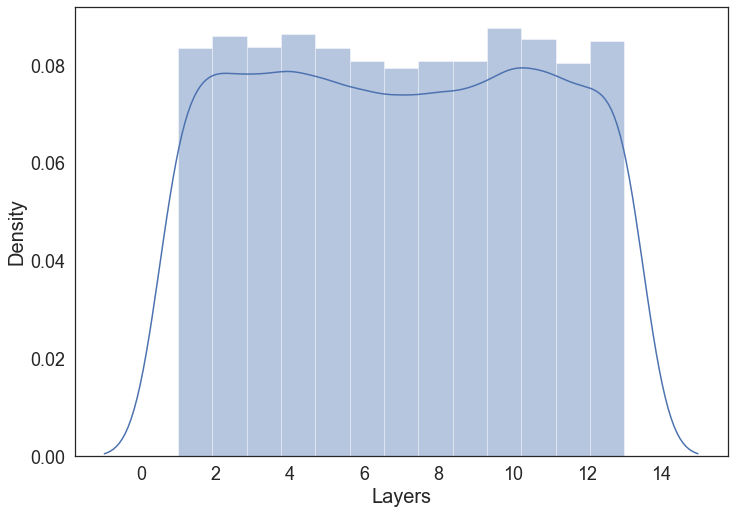}\label{fig:7l_distribution}}\\
    \subfloat[Train 75\% (10 Layers) of the model ] {\includegraphics[width=0.55\textwidth]{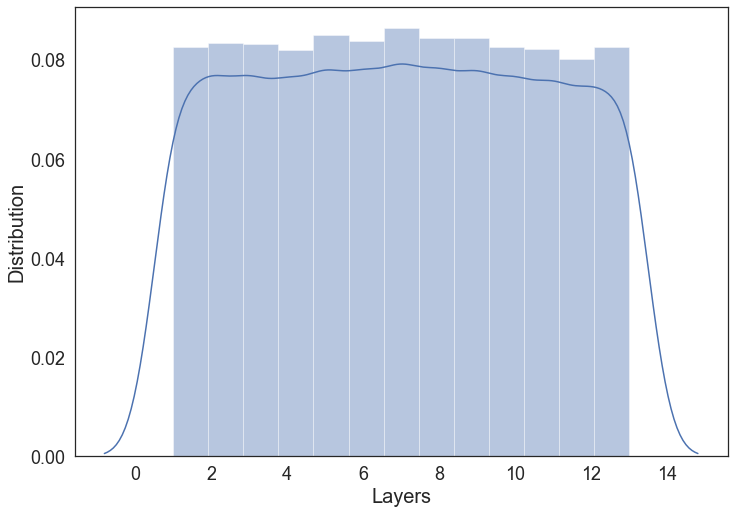}\label{fig:10l_distribution}}
    \caption{VGG16 layers distribution across 10 clients during 100 training rounds using different parts of the model}
    \label{fig:layersDistribution}
\end{figure*}

As the number of clients participating in a training round increases, the probability of engaging all layers of a model in the global training process also rises. Simultaneously, the volume of training data expands. This heightened participation and increased training data elevate the likelihood of training all model layers over each client's data as multiple rounds progress. Consequently, this phenomenon significantly influences the model's convergence.This effect is particularly noticeable when we randomly select different trainable layers of the model ($25\%$, $50\%$, and $75\%$) for training. During the training process, we noticed that each client has the opportunity to train every layer of the model at least once. Moreover, the distribution of layers among all clients is equitable, ensuring a balanced contribution to the training process, as demonstrated in Figure~\ref{fig:layersDistribution}. The equal distribution of model layers has been used across all clients with different training settings (i.e. $4$, $7$, and $10$ layers).

\subsubsection{The impact of scaling the number of clients (edge nodes) on the model accuracy}
We conducted two more experiments to estimate the impact by using different settings, such as the number of clients, and the number of trainable layers. These experiments aimed to determine how these factors affect model accuracy while keeping the amount of data fixed.
\begin{itemize}

\item \textbf{Exp. 1}: In this experiment, we partitioned the dataset among $10$ distinct clients to train the entire model, which consisted of $14$ trainable layers, within the context of FL settings.

\item \textbf{Exp. 2}:  In this particular experiment, we divided the dataset into $20$ partitions, distributed across $20$ clients. The objective was to train $7$ trainable layers, which were randomly selected during each training round in the FL context. This corresponds to training half of the entire model in each round.

\end{itemize}

\begin{figure}[!h]
\centering
\includegraphics[width=0.9\textwidth]{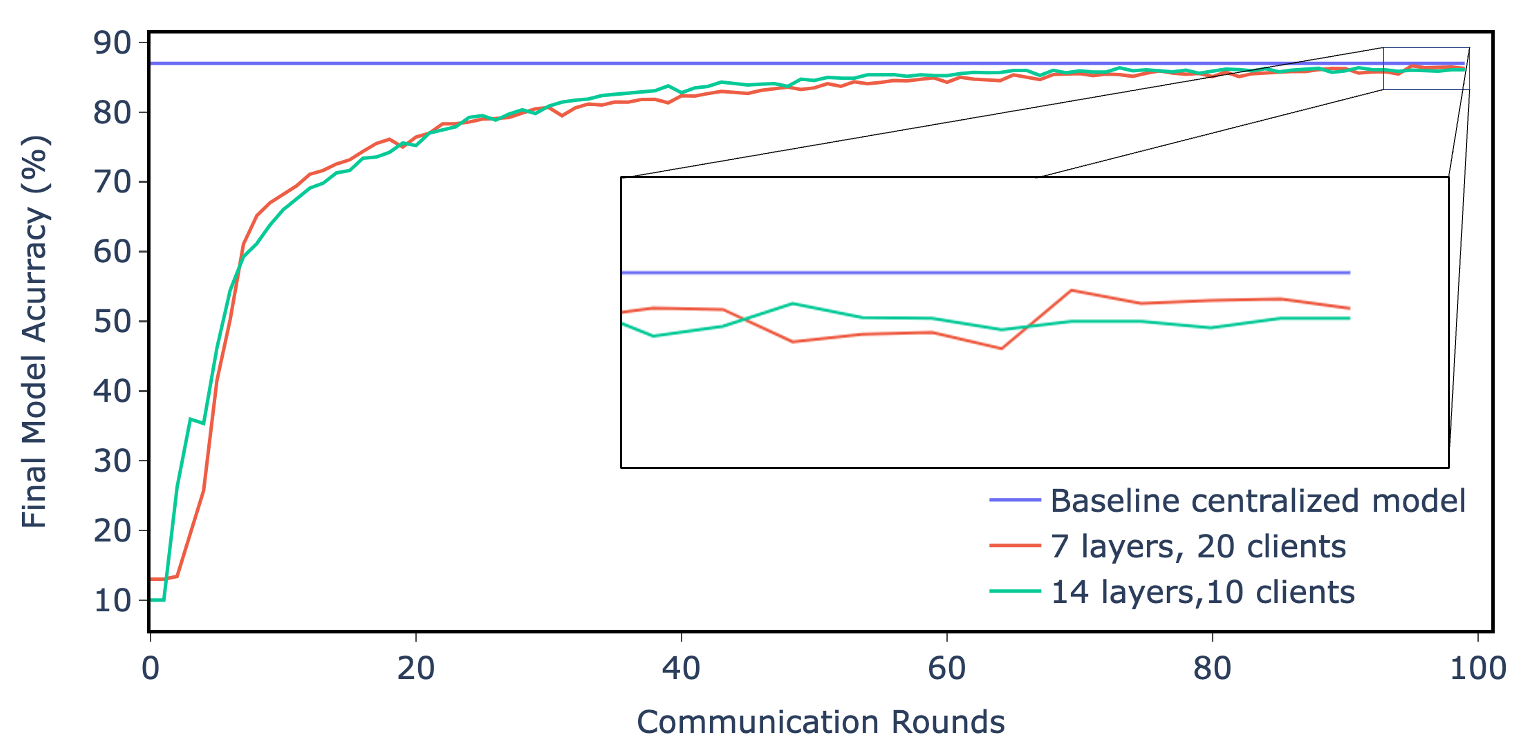}
\caption{ Comparing the impact of reducing the number of trainable layers to half ($7$ layers) while scaling the number of clients ($20$ layers) to double the number of clients ($10$ layers) used to train the whole model ($14$ layers). This setting change was carried out while maintaining the same amount of CIFAR-10 data for both scenarios. The objective was to evaluate how these modifications influenced the global model's accuracy.
}
\label{fig:AccDoubCli}
\centering
\end{figure}

Figure \ref{fig:AccDoubCli} shows the impact of scaling the number of clients, using different numbers of trainable layers of the VGG16 model while keeping the data amount fixed. According to the results, both experimental models demonstrated high accuracy (86.08\% and 86.28\%) in comparison to the centralized model's accuracy, with only a minor difference of approximately  0.92\% and 0.72\%, respectively.

We found that using more clients (with fewer resources) and fewer trainable layers can achieve the same model performance as training the entire model with fewer nodes. This was particularly evident in the last $20$ rounds, where double the number of nodes were used to train $7$ layers. That leads to the conclusion, that with more clients, each layer had more opportunities to be trained at least once per communication round, resulting in better accuracy.

\begin{figure}[!h]
\centering
\includegraphics[width=0.9\textwidth]{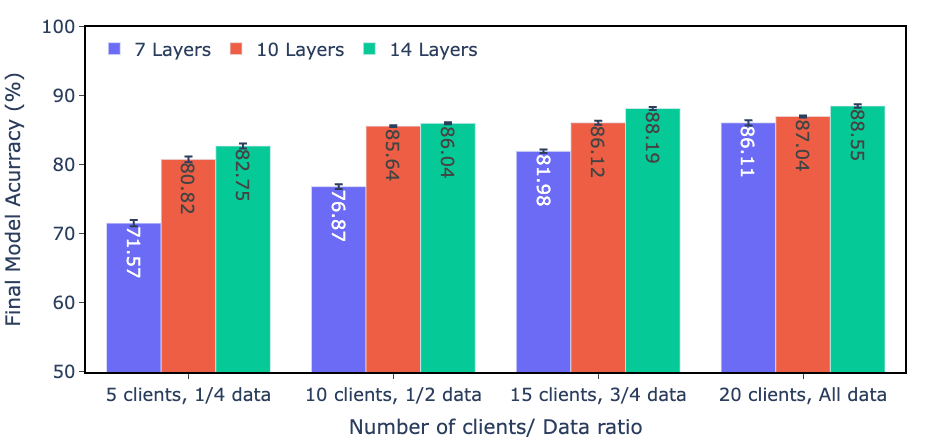}
\caption{Show the impact of scaling the number of clients, adjusting the number of model layers, and increasing the dataset size on the global model's performance. Training approximately 75\% of the model yielded a high accuracy level, closely approaching the fully trained model, with only a tiny performance gap. Furthermore, training 50\% (half) of the model still achieved a high level of accuracy, with negligible differences observed in the model's performance as the number of clients increased, as demonstrated in the case involving 20 clients.
}
\label{fig:AccScalClients}
\end{figure}

Another experiment has been conducted to evaluate how scaling the number of clients, data ratio and trainable layers affected model accuracy. As depicted in Figure \ref{fig:AccScalClients}, the bar chart demonstrates a consistent enhancement in model accuracy as the number of clients are increased across all training settings. Specifically, when training the model with seven layers, scaling the contributors from $5$ to $20$, and expanding the training data from a quarter to all the data, an accuracy gain of approximately $15\%$ was observed. The same scenario was followed for $10$ and $14$ trainable layers.

Training the model with different numbers of layers using varying numbers of clients had minimal impact on model performance, even with an increased amount of training data. Comparing the accuracy achieved using 20 clients to train 7 and 10 layers of the model revealed only a borderline difference of approximately 1\% in accuracy, as shown in Figure \ref{fig:AccScalClients}. A slightly more significant gap was observed when comparing the accuracy obtained by training 7 and 14 layers of the model using 20 clients, with a difference of around 2\%.

These findings indicate that training a model with fewer layers can be a more resource-efficient alternative without significantly compromising accuracy. The advantage is that fewer layers requires less computational resources client side. This is particularly important while dealing with restricted devices or limited computational capabilities.

The results also highlight the scalability of the approach, as increasing the number of clients can compensate for the reduced model depth, resulting in comparable accuracy to training with more clients and fewer layers.

\begin{figure}[!h]
\centering
\includegraphics[width=0.9\textwidth]{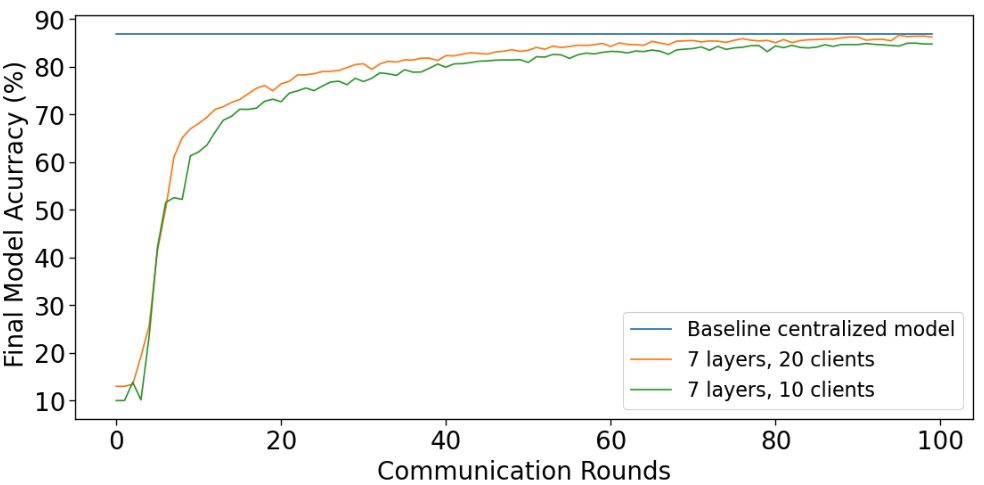}
\caption{The impact of training a consistent number of model layers (7) over a fixed amount of data while scaling the number of clients on model performance. Notably, the model's accuracy demonstrated improvement as the number of clients increased, ultimately outperforming the model trained with fewer clients.
}
\label{fig:scaleClientFixData}
\centering
\end{figure}

Moreover, we evaluate the effectiveness of our approach by increasing the number of clients while keeping the data amount constant to train 7 layers. As shown in Figure \ref{fig:scaleClientFixData}, Despite employing the same model architecture and total data size, we have observed similar performance when using either 20 or 10 clients. It is important to note that we divided the dataset into 10 and 20 partitions, which were assigned to 10 and 20 clients, respectively. By increasing the number of clients involved in constructing the global model, we noticed an improvement in the model's performance. Therefore, our findings indicate that the global model generated by 20 clients outperformed the one built by 10 clients.
Furthermore, we observed that the model trained with 20 clients achieved good accuracy, albeit slightly different from the baseline. This can be attributed to the model's capability to learn hidden patterns from data with sufficient samples more efficiently, even when the client has a smaller sample size. These findings highlight the significance of involving more contributors (clients) and selecting an appropriate number of trainable layers in the FL settings, as they substantially impact the model's performance. Additionally, this outcome allows for a more precise estimation of the training budget and optimizing the utilization of available resources.

\subsubsection{Training time}
\label{sec:trainingTime}

In the previous sections, we demonstrated the ability to maintain accuracy while freezing parts of the model. Nonetheless, it is crucial to consider the effect on training time, mainly when dealing with larger models. Nevertheless, our approach demonstrates its effectiveness in accelerating the training process. This is achieved by distributing sub-layers of the entire model among the clients to be trained over clients' local data within the FL network, ultimately reducing training time. 

\begin{figure}[!h]
\centering
\includegraphics[width=0.9\textwidth]{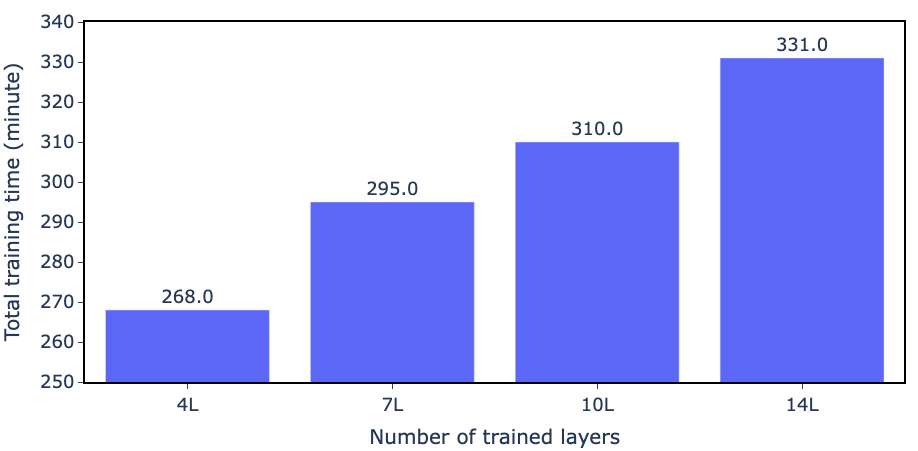}
\caption{Total time required for 10 clients to complete 100 global rounds when training different parts of VGG16 model over CIFAR-10 dataset.}
\label{fig:totalTraining10C100R}
\centering
\end{figure}

Figure~\ref{fig:totalTraining10C100R} illustrates the total training time (in minutes) required by 10 clients to complete 100 training rounds. Training the entire model takes approximately 331 minutes while training 75\% of the model saves approximately 21 minutes (a 7\% reduction in time). There is a significant difference of 63 minutes between training 4 layers (25\% of the model) compared to training 14 layers. By training approximately 50\% of the model, we can save around 36 minutes (a 10\% reduction compared to training the entire model) while still maintaining accuracy (refer to Figure~\ref{fig:modelAccuracy}). Take into account that the observed time reduction is relatively low, influenced by various factors such as the computational resources of individual clients, the size of the local datasets, and the presence of straggler clients, making the training round longer.

\begin{figure}[!h]
\centering
\includegraphics[width=0.9\textwidth]{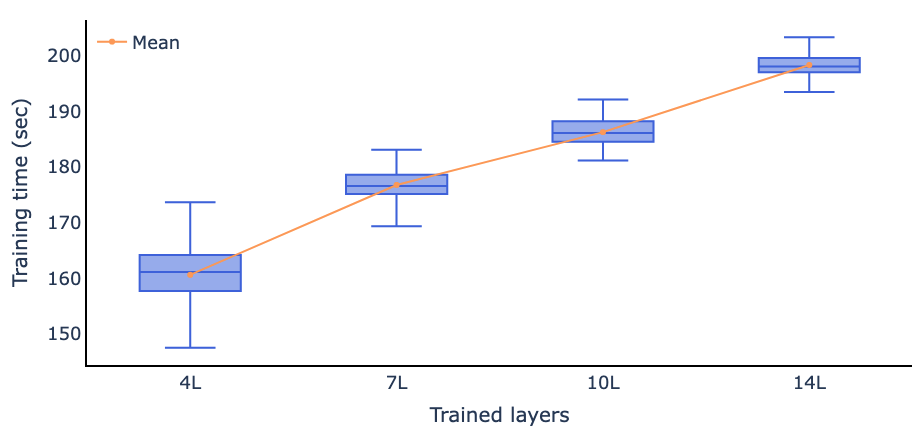}
\caption{The time spent by one client to train the VGG16 model over CIFAR-10 dataset using  different trainable layers (cost per client per global communication round).
}

\label{fig:totalTraining1C100R}
\centering
\end{figure}

Additionally, Figure~\ref{fig:totalTraining1C100R} demonstrates the time required for a single client to complete 100 training rounds. As the number of trained layers increases, the training time grows linearly, impacting computational resources in terms of cost and availability.

Analyzing the training time across various rounds to achieve a specific accuracy level is crucial. Table \ref{tab:timeAnalys_rounds} illustrates the time required to train 4, 7, 10 and 14 layers, respectively, and the corresponding accuracy obtained after 40, 60, 80 and 100 training rounds.   

Training 4 layers selected randomly from the complete model achieves 70.54\% accuracy within 40 rounds, which requires 104 minutes. Notably, increasing the number of rounds by 20 each time consistently requires around 53 minutes, demonstrating a constant trend. Furthermore, a substantial enhancement in model performance is observed, obtaining 79.02\% accuracy after 100 rounds. This pattern is consistent across different layers. Variations were observed when comparing the execution time for training different layers within the same communication rounds category. For instance, the time difference between training 7 and 10 layers for 40 rounds is 6 minutes, and for 60 rounds is 9 minutes. However, the accuracy difference among different layers with the same number of rounds remained relatively close. Comparing a fully trained model (14 layers) after 100 rounds with 7 and 10 layers shows an insignificant accuracy difference, confirming that training a portion of the model while maintaining good accuracy.

\begin{table*}[!htb]
\caption{The time required (in minute) to achieve a certain level of accuracy (Acc.) during various communication rounds to train the VGG16 model by 10 clients in the FL context.}
\label{tab:timeAnalys_rounds}
\begin{center}
\begin{tabular}{lllllllll}
\cmidrule{2-9}
& \multicolumn{2}{c}{\textbf{40 rounds}}    & \multicolumn{2}{c}{\textbf{60 rounds}}    & \multicolumn{2}{c}{\textbf{80 rounds}}                     & \multicolumn{2}{c}{\textbf{100 rounds}}                    \\ 
\cmidrule{1-9}

\multicolumn{1}{l|}{\textbf{Number of Layers}} & \multicolumn{1}{l}{\textbf{Time}} & \multicolumn{1}{l|}{\textbf{Acc.}} & \multicolumn{1}{l}{\textbf{Time}} & \multicolumn{1}{l|}{\textbf{Acc.}} & \multicolumn{1}{l}{\textbf{Time}} & \multicolumn{1}{l|}{\textbf{Acc.}} & \multicolumn{1}{l}{\textbf{Time}} & \multicolumn{1}{l}{\textbf{Acc.}} \\  \cmidrule{1-9}

\multicolumn{1}{l|}{\textbf{4 Layers}}         & \multicolumn{1}{l}{107}           & \multicolumn{1}{l|}{70.54\% }       & \multicolumn{1}{l}{161}           & \multicolumn{1}{l|}{75.95\%}        & \multicolumn{1}{l}{214}           & \multicolumn{1}{l|}{77.72\% }       & \multicolumn{1}{l}{268}           & \multicolumn{1}{l}{79.02\%}        \\ 
\multicolumn{1}{l|}{\textbf{7 Layers}}         & \multicolumn{1}{l}{118}           & \multicolumn{1}{l|}{79.95\%}        & \multicolumn{1}{l}{177}           & \multicolumn{1}{l|}{83.26\%}        & \multicolumn{1}{l}{236}           & \multicolumn{1}{l|}{84.4\% }        & \multicolumn{1}{l}{295}           & \multicolumn{1}{l}{84.75\%}        \\ 
\multicolumn{1}{l|}{\textbf{10 Layers}}        & \multicolumn{1}{l}{124}           & \multicolumn{1}{l|}{81.73\%}        & \multicolumn{1}{l}{186}           & \multicolumn{1}{l|}{83.82\%}        & \multicolumn{1}{l}{248}           & \multicolumn{1}{l|}{85.08\% }       & \multicolumn{1}{l}{310}           & \multicolumn{1}{l}{85.68\% }       \\ 
\multicolumn{1}{l|}{\textbf{14 Layers}}        & \multicolumn{1}{l}{132}           & \multicolumn{1}{l|}{82.79\% }       & \multicolumn{1}{l}{199}           & \multicolumn{1}{l|}{85.25\% }       & \multicolumn{1}{l}{265}           & \multicolumn{1}{l|}{85.88\% }       & \multicolumn{1}{l}{331}           & \multicolumn{1}{l}{86.08\%}        \\ 
\hline

\end{tabular}
\end{center}
\end{table*}

\subsubsection{Transferred data size and number of trainable parameters }
\label{sec:transData}
The size of transferred parameters (weights) in the FL settings naturally depends on the number of trainable layers in the neural network. In addition to the benefits mentioned above, layer sub-selection can potentially reduce the amount of resources required on the client side. This can include computing power, storage space, and network capacity needed for communication


\begin{table}[!htb]
\caption{The average transferred data size for different numbers of trained layers during a  communication round with 10 participating clients}
\label{tab:parameters_dataSize}
\begin{center}
\scalebox{.86}{
\begin{tabular}{l|l|l|l|l|}
\cline{2-5}
                                                          & \textbf{4 Layers} & \textbf{7 Layers} & \textbf{10 Layers} & \textbf{14 Layers} \\ \hline
\multicolumn{1}{|l|}{\textbf{No. of training parameters}} & 34.88 M           & 67.92 M           & 101.3 M            & 147.2 M            \\ \hline
\multicolumn{1}{|l|}{\textbf{Transferred data size}}      & 133.1 MB          & 259.1 MB          & 386.5 MB           & 561.6 MB           \\ \hline
\end{tabular}
}
\end{center}
\end{table}


Table~\ref{tab:parameters_dataSize} presents the number of trainable parameters and transferred data size for different training settings over 100 training rounds with 10 clients, highlighting the linear correlation between the number of trainable layers and serialized data size.

Additionally, we observed a significant time difference (see Figure~\ref{fig:totalTraining10C100R}) between training a model with 4 layers (268 minutes) and 14 layers (331 minutes), with a difference of 63 minutes. When considering factors such as model accuracy (refer to Figure~\ref{fig:modelAccuracy}), transferred updated gradients, number of trainable parameters (as shown in Table~\ref{tab:parameters_dataSize}), and training time (see Figure~\ref{fig:totalTraining10C100R}), 
these effects will be discussed in more detail in the upcoming section.

\subsubsection{Resources utilization based on VM's} \label{sec:resour1}

In this section we empirically study practical resource constraints by varying the computational resources available clients side by using different VM flavors in the SNIC Science Cloud. Table \ref{clientsFlavour} reports the measured percentage of CPU and RAM utilization for local training when varying the number of trainable layers. We conducted experiments starting from the ssc.xsmall flavor, which simulates devices with restricted resources, and gradually scaled up the client resources to the ssc.xlarge flavor. As shown in Table \ref{clientsFlavour}, using the ssc.xsmall flavor allowed us to train only 4 layers of the VGG16 model, utilizing both CPU and RAM fully, and requiring 1119.49 seconds. Due to resource limitations, the client cannot train additional layers using the available resources. However, by scaling up to the ssc.small flavor, we were able to train up to 10 layers without any issues. Training the entire model required the essential resource of the ssc.small.highcpu flavor (2 VCPUs, 2 RAM). When training 4 layers, 90\% of the CPU and 64.64\% of the RAM were allocated, with a training time of 535.24 seconds compared to the ssc.xsmall flavor. Training half of the model resulted in a 4\% increase in CPU and RAM utilization, slowing down the training process by approximately 11 seconds. Comparing the local training process for 14 and 4 layers, we observed a significant resource utilization gap (CPU: 8\%, RAM: 12.14\%) and an increase in training time by 84.41 seconds, highlighting the need for more computational power to perform the task.  

To expedite the training process for a single client, it often demands an increase in computational resources, such as CPU and RAM. However, this can lead to higher overall training costs and limitations in scaling up these resources, potentially resulting in a single point of failure. 
As a solution, scaling out the number of clients with lower resources can mitigate these challenges. Training the model partially while increasing the number of clients, we've shown in Figure \ref{fig:AccScalClients} that it doesn't adversely affect the model's accuracy. Consequently, this strategy allows for the model training process to involve constrained devices like Jetson Nano or VM's with limited configurations (e.g., ssc.small). This approach enables the establishment of a cluster incorporating restricted nodes to train the ML model in FL setting. The primary goal is to reduce training costs and enhance resilience against failures.

\begin{table*}[h]
\caption{The local training cost in terms of time (seconds), CPU, and RAM per round may vary for different clients with different flavors deployed on the SNIC Science Cloud. \textbf{Flavors:} \textit{ssc.xsmall (1 VCPU, 1 GB RAM),ssc.small (1 VCPU, 2 GB RAM), ssc.small.highcpu (2 VCPU, 2 GB RAM), ssc.medium (2 VCPU, 4 GB RAM), ssc.medium.highcpu (4 VCPU, 4 GB RAM), ssc.large (4 VCPU, 8 GB RAM), ssc.xlarge (8 VCPU, 16 GB RAM)}}
\label{clientsFlavour}
\scalebox{.53}{
\begin{tabular*}{\textheight}{@{\extracolsep\fill}lllllllllllllll}
\cmidrule{1-13}
\multirow{3}{*}{\textbf{Flavour}} & \multicolumn{4}{@{}c@{}}{\textbf{Training time (Sec.)}} & \multicolumn{8}{@{}c@{}}{\textbf{Resources Consumption}}  \\\cmidrule{2-5}\cmidrule{6-13}

& \multirow{2}{*}{\textbf{4 Layers}} &\multirow{2}{*}{\textbf{7 Layers}} & \multirow{2}{*}{\textbf{10 Layers}} & \multirow{2}{*}{\textbf{14 Layers}} & \multicolumn{2}{c}{\textbf{4 Layers}} & \multicolumn{2}{c}{\textbf{7 Layers}}  & \multicolumn{2}{c}{\textbf{10 Layers}} & \multicolumn{2}{c}{\textbf{14 Layers}} \\ \cmidrule{6-13} 
& &   & &  & CPU & RAM (MB) & \multicolumn{1}{|l}{CPU} & RAM (MB)& \multicolumn{1}{|l}{CPU} & RAM (MB)&\multicolumn{1}{|l}{CPU}& RAM (MB) \\ \cmidrule{1-13}

\rowcolor{LightCyan}
                                  
 ssc.xsmall   & \multicolumn{1}{l|}{1119.49}  & \multicolumn{1}{l|}{-}  & \multicolumn{1}{l|}{-} &  \multicolumn{1}{l||}{-}                                               & \multicolumn{1}{l|}{99\%}  & \multicolumn{1}{l|}{988 (95.70\%)}     & \multicolumn{1}{l|}{-}  & \multicolumn{1}{l|}{-}     & \multicolumn{1}{l|}{-}  & \multicolumn{1}{l|}{-}     & \multicolumn{1}{l|}{-}  & \multicolumn{1}{l|}{-}   \\

\rowcolor{LightCyan}

ssc.small                   & \multicolumn{1}{l|}{894.0}                             & \multicolumn{1}{l|}{954.0}                             & \multicolumn{1}{l|}{1086.0}                              & \multicolumn{1}{l||}{-}                                                    & \multicolumn{1}{l|}{99\%}  & \multicolumn{1}{l|}{1312 (66.21\%)}     & \multicolumn{1}{l|}{99\%}  & \multicolumn{1}{l|}{1490 (72.75\%)}     & \multicolumn{1}{l|}{99\%}  & \multicolumn{1}{l|}{1758 (85.83\%)}     & \multicolumn{1}{l|}{-}  & \multicolumn{1}{l|}{-}                        \\

ssc.small.highcpu                & \multicolumn{1}{l|}{535.24}                             & \multicolumn{1}{l|}{546.23}                             & \multicolumn{1}{l|}{604.65}                              &  \multicolumn{1}{l||}{619.65}                                                   & \multicolumn{1}{l|}{90\%}  & \multicolumn{1}{l|}{1284 (64.64\%)}     & \multicolumn{1}{l|}{94\%}  & \multicolumn{1}{l|}{1373 (69.10\%)}     & \multicolumn{1}{l|}{96\%}  & \multicolumn{1}{l|}{1438 (72.39\%)}     & \multicolumn{1}{l|}{98\%}  &    \multicolumn{1}{l|}{1525 (76.78\%)}                     \\

ssc.medium                      & \multicolumn{1}{l|}{467.16}                             & \multicolumn{1}{l|}{533.75}                             & \multicolumn{1}{l|}{555.92}                              &   \multicolumn{1}{l||}{582.22}                                                  & \multicolumn{1}{l|}{80\%}  & \multicolumn{1}{l|}{1410 (35.85\%)}     & \multicolumn{1}{l|}{87\%}  & \multicolumn{1}{l|}{1486 (37.77\%)}     & \multicolumn{1}{l|}{90\%}  & \multicolumn{1}{l|}{1559 (39.64\%)}     & \multicolumn{1}{l|}{93\%}  &                \multicolumn{1}{l|}{1678 (42.63\%)   }         \\

ssc.medium.highcpu               & \multicolumn{1}{l|}{280.87}                             & \multicolumn{1}{l|}{310.19}                             & \multicolumn{1}{l|}{323.15}                              &     \multicolumn{1}{l||}{357.6 }                                                & \multicolumn{1}{l|}{71\%}  & \multicolumn{1}{l|}{1519 (38.63\%)}     & \multicolumn{1}{l|}{77\%}  & \multicolumn{1}{l|}{1660 (42.20\%)}     & \multicolumn{1}{l|}{82\%}  & \multicolumn{1}{l|}{1662 (42.20\%)}     & \multicolumn{1}{l|}{88\%}  &                 \multicolumn{1}{l|}{1698 (43.18\%)    }         \\

ssc.large                       & \multicolumn{1}{l|}{270.26}                             & \multicolumn{1}{l|}{303.13}                             & \multicolumn{1}{l|}{315.37}                              &     \multicolumn{1}{l||}{342.66}                                                & \multicolumn{1}{l|}{68\%}  & \multicolumn{1}{l|}{1541 (19.38\%)}     & \multicolumn{1}{l|}{76\%}  & \multicolumn{1}{l|}{1602 (20.13\%)}     & \multicolumn{1}{l|}{79\%}  & \multicolumn{1}{l|}{1661 (20.86\%)}     & \multicolumn{1}{l|}{86\%}  &           \multicolumn{1}{l|}{1723 (21.64\%)    }                 \\

ssc.xlarge                       & \multicolumn{1}{l|}{160.24}                             & \multicolumn{1}{l|}{180.08}                             & \multicolumn{1}{l|}{187.04}                              &        \multicolumn{1}{l||}{195.86 }                                            & \multicolumn{1}{l|}{55\%}  & \multicolumn{1}{l|}{1795 (11.22\%)}    & \multicolumn{1}{l|}{60\%}  & \multicolumn{1}{l|}{1934 (12.09\%)}    & \multicolumn{1}{l|}{64\%}  & \multicolumn{1}{l|}{1988 (12.42\%)}    & \multicolumn{1}{l|}{78\%}  &              \multicolumn{1}{l|}{2059 (12.86\%)    }           \\ 
\cmidrule{1-13}
\end{tabular*}
}

\end{table*}


\subsubsection{Resources utilization based on Jetson Nano}\label{sec:resour2}

 We conducted an additional experiment using an actual restricted device, the Jetson Nano 2GB kit \footnote{\url{https://developer.nvidia.com/embedded/jetson-nano-2gb-developer-kit}}. This experiment aimed to observe the behaviour of the device in terms of resource utilization during the local model training process for different subsets of layers \footnote{\url{https://github.com/saadiabadi/Jetson-Nano-Setup-Fedn.git}}. We utilized a lighter version of the VGG16 model by reducing the dimensions of the previous model's layers by half and setting the batch size to 4. This adjustment aimed to avoid out-of-memory issues.

Table \ref{jetson-nano} reports the training time and resource consumption per training round for different trainable layers on the Jetson Nano. It can be observed that training 4 layers requires 191.4 seconds per round and demands 46.55\% and 88.61\% of the CPU and RAM, respectively. These resource demands are still manageable for the device. However, as we increase the number of trained layers, these values increase accordingly. Comparing the training time for 4 and 10 layers, we noticed a difference of approximately 66 seconds, which can be significant in critical scenarios. Furthermore, training 10 layers resulted in a 4\% and 7\%  increase in both CPU and RAM consumption, respectively.

\begin{table}[!htb]
\scriptsize
\caption{The local training cost in terms of time (Seconds), CPU and RAM per round using Jetson Nano 2GB client (Quad-core ARM CPU, 2GB RAM)}
\label{jetson-nano}
\begin{center}
\begin{tabular}{|cccccccc|}
\hline \multicolumn{8}{|c|}{\textbf{Training time (Sec.)}}                                                                                                                                                                                                          \\ \hline 
                                  \multicolumn{2}{|c|}{\textbf{4 Layers}}                                      & \multicolumn{2}{c|}{\textbf{7 Layers}}                                      & \multicolumn{2}{c|}{\textbf{10 Layers}}                                     & \multicolumn{2}{c|}{\textbf{14 Layers}}      
                                  \\ \hline \multicolumn{2}{|c|}{191.4}                                         & \multicolumn{2}{c|}{235.6}                                         & \multicolumn{2}{c|}{257.8}                                         & \multicolumn{2}{c|}{\cellcolor{blue!25}-}

\\ \hline \multicolumn{8}{|c|}{\textbf{Resources Consumption}}                                                                                                                                                                                                         \\ \hline
\multicolumn{1}{|c|}{\textbf{CPU}}     & \multicolumn{1}{c|}{\textbf{RAM (MB)}}       & \multicolumn{1}{c|}{\textbf{CPU}}     & \multicolumn{1}{c|}{\textbf{RAM (MB)}}       & \multicolumn{1}{c|}{\textbf{CPU}}     & \multicolumn{1}{c|}{\textbf{RAM (MB)}}       & \multicolumn{1}{c|}{\textbf{CPU}} & \textbf{RAM (MB)} 

\\ \hline \multicolumn{1}{|c|}{46.55\%} & \multicolumn{1}{c|}{1747 (88.61\%)} & \multicolumn{1}{c|}{48.23\%} & \multicolumn{1}{c|}{1821 (92.38\%)} & \multicolumn{1}{c|}{50.63\%} & \multicolumn{1}{c|}{1881 (95.68\%)} & \multicolumn{1}{c|}{\cellcolor{blue!25}-}    & {\cellcolor{blue!25}-} \\ \hline
                                 
\end{tabular}
\end{center}
\end{table}

However, during the training process of the entire model, the Jetson Nano crashed (cannot finish the training) due to a lack of memory. Finally, this experiment on the Jetson Nano device further supports the benefits of training the model partially on restricted devices, as it allows for efficient resource utilization and avoids memory limitations.

\subsection{Limitations of the proposed approach}

The proposed approach can be implemented in any FL framework with consistent behaviour. Integrating differential privacy (DP) into our technique and comparing it with vanilla FL with DP may impact the results, but our approach maintains its fundamental characteristics. Adding DP introduces additional thresholds to both, significantly influencing training time and model accuracy.

However, it's important to acknowledge certain limitations in the proposed approach:
\begin{enumerate}

\item The number of selected layers remains fixed during the FL training process, potentially affecting client performance and execution time in local training. This constraint could be mitigated by dynamically selecting an optimal number of layers each round based on the ideal resources for each client.

\item With the emergence of large language models (LLM), the approach might require additional engineering to handle the sophistication of such complex models.

\item Each client's sequential training of layers represents a limitation that could be addressed through more sophisticated strategies in future iterations.

\end{enumerate}

\section{Conclusion} \label{sec:conc}
This paper introduces a novel approach for training DL models in the FL setting, aiming to efficiently utilize edge node resources and reduce network workload. The proposed approach trains a specific number of the model's layers, randomly selected every training round, and freezes the remaining layers. This approach can enable IoT devices with restricted resources (as exemplified here with the Jetson Nano) to participate in training larger models.

The approach was evaluated for three tasks: sentiment analysis using the IMDB dataset, object detection using the CIFAR-10 dataset, and human activity recognition using the CASA dataset, each with different model architectures. The experimental results demonstrate that training only a part of the model in the FL setting has a significant impact on resource utilization, communication, training budget, and model performance.  
Furthermore, increasing the number of contributors also considerably affects model performance while keeping the amount of data fixed as we demonstrated in the results.

Overall, the study demonstrates the potential of our approach to train DL models in a FL setting efficiently, enabling participation from a diverse range of devices and clients without sacrificing model performance.

In future work, the authors plan to investigate strategies for selecting layers based on the available resources of each client and the number of trainable parameters per layer, considering the expected heterogeneity in real-life use cases. Additionally, exploring measures of each layer's importance for model improvement could be an interesting avenue for further research.



\section*{Acknowledgement}
This work was funded by the eSSENCE strategic collaboration on eScience (Alawadi, Ait-Mlouk, Toor, and Hellander), and supported by Blekinge Institute of Technology (BTH). The authors also would like to thank SNIC for providing cloud resources.

\bibliographystyle{unsrtnat}

\bibliography{sample-base}
\end{document}